# Information Density as a Factor for Variation in the Embedding of Relative Clauses[*]

Augustin Speyer (000-0003-1027-2635), Robin Lemke (0000-0003-2964-7396)

Saarland University, Saarbrücken, Germany

`a.speyer@mx.uni-saarland.de`

**Abstract.** In German, relative clauses can be positioned in-situ or extraposed. A potential factor for the variation might be information density. In this study, this hypothesis is tested with a corpus of 17[th] century German funeral sermons. For each referent in the relative clauses and their matrix clauses, the attention state was determined (first calculation). In a second calculation, for each word the surprisal values were determined, using a bi-gram language model. In a third calculation, the surprisal values were accommodated as to whether it is the first occurrence of the word in question or not. All three calculations pointed in the same direction: With in-situ relative clauses, the rate of new referents was lower and the average surprisal values were lower, especially the accommodated surprisal values, than with extraposed relative clauses. This indicated that information density is a factor governing the choice between in-situ and extraposed relative clauses. The study also sheds light on the intrinsic relationship between the information theoretic concept of information density and information structural concepts such as givenness which are used under a more linguistic perspective.

**Keywords:** Relative Clause · Extraposition · Information Density · Givenness · Language Models

## 1 Introduction

### 1.1 The Phenomena

German displays several cases of seemingly free variation of syntactic variants, most notably for expressing complex propositions. These cases include

- the realization of adverbial clauses as subordinate clauses embedded in their matrix clause (1a) or not embedded (1b); non-embeddedness being visible e.g. in the usage of verb-second syntax,
- the realization of argument clauses as non-finite clauses (2a), as subordinate claues with features of embeddedness (e.g. verb-final syntax, 2b), or as subordinate clauses with features of non-embeddedness (e.g. verb-second syntax, lack of complementizer, 2c),
- the positioning of relative clauses adjacent to their head noun (3a) or in the so-called post-field, i.e. after the position of the non-finite parts of the verb form (3b)
- the positioning of non-clausal constituents within the clause proper in the middle field (4a), that is: before the position of the non-finite parts of the verb form, or in the post-field (4b).

(1) a. Jörg kam zu spät, weil sein Auto nicht ansprang.
Jörg came too late because his car not started
b. Jörg kam zu spät. Denn sein Auto sprang nicht an.
Jörg came too late. For his car started not ptc.
'Jörg came late, because his car did not start.'

(2) a. Julia versprach, die Klausuren zu korrigieren
Julia promised the exams to grade
b. Julia versprach, dass sie die Klausuren korrigieren werde.
Julia promised that she the exams grade would
c. Julia versprach, sie werde die Klausuren korrigieren.
Julia promised she would the exams grade

[*] We wish to thank our colleagues, especially Bernd Möbius, Petra Schumacher, and Elke Teich, for their valuable input, and the research assistants of one of us (Speyer), Hannah Koob, Yannick Schramm, and Sophia Voigtmann, for their help in preparing the data and calculating the models.

'Julia promised that she would grade the exams / Julia promised to grade the exams'

(3) a. Philipp hat den Studenten, der das Buch ausgeliehen hat, gesehen.
       Philipp has the student who the book borrowed has seen
    b. Philipp hat den Studenten gesehen, der das Buch ausgeliehen hat.
       Philipp has the student seen who the book borrowed has
       'Philipp saw the student who had borrowed the book.'

(4) a. Wir haben zunächst die Bushaltestelle wegen der Rückfahrt gesucht.
       we have first the bus-stop for the journey-back searched
    b. Wir haben zunächst die Bushaltestelle gesucht wegen der Rückfahrt.
       we have first the bus-stop searched for the journey-back
       'First we looked for the bus stop because of the journey back.'

What it the factor that triggers this variation? It might have to do with the distribution of information: All of these variants have in common that either all of the information is mashed together, so to speak, in one complex syntactic unit (this goes for the variants in (1a,2a,b,3a,4a) or that the information is distributed over several syntactic units (1b,2c,3b,4b).[1] Let us refer to the variants displayed in (1a) etc. as bundling variants, and to the variants displayed in (1b) etc. as outsourcing variants. The case illustrated in (2) shows that there is not a dichotomy between bundling and outsourcing but rather a continuum.

Some of these phenomena have been linked to information structural concepts such as e.g. givenness in the literature.[2] Very roughly speaking, the outsourcing variants have been found to correspond to information that is not given and/or focussed, while the bundling variants correspond to given information.

The same kinds of variation were visible also in earlier stages of German (e.g. Speyer 2015, 2016, to appear). However, the relative frequency of outsourcing versus bundling variants changed over time. To give two examples: The ratio of fully integrated causal clauses rose from c. 40% in the 15th century to c. 53% in the 16th century (Speyer 2015), the rate of clauses without postfield from 68% in 1500 to 81% in 1700 (Schildt 1976). So obviously some change occurred in the course of German language history which has to do with the distribution of information.

## 1.2 The Hypothesis

Our main research question is: What determines the choice between the bundling and the outsourcing variants? Our hypothesis is that if a clause contains information which is hard to process language producers tend to opt for outsourcing variants. We assume a model of language generation in the tradition of Levelt 1989 (see also e.g. Schumacher & Hung 2012) in which propositions – which correspond roughly to clauses – are planned in one chunk. If it becomes clear that the proposition which the producer is about to encode surpasses a certain threshold of complexity so that it is difficult to compute, the producer assigns parts of it to another processing unit.[3] If the proposition contains subordinated propositions, the natural choice of material to outsource is the content of the subordinate propositions. The result are subordinate clauses that are not integrated. If the proposition does not contain subordinated propositions, some constituent(s) is/are singled out and assigned to the postfield which counts as a separate processing unit (e.g. Auer 1991, Hoberg 1997).An important question in this context is what determines whether a piece of information is hard or easy to compute. There are several answers to that. Working in an information structural framework, one might assume that the computability of an expression depends on the information structural status that the entity to which the expression refers possesses. If an element represents given information, or counts as the topic, its computation is easier than if it represents e.g. new information. This connection between accessibility and computability, understood as activation, is at the basis of much work on salience (e.g. Gernsbacher 1989, Kaiser & Trueswell 2011, de la Fuente & Hemforth 2013). A problem with information structural approaches is that these information structural statuses are hard to quantify.

---

[1] Fully integrated clauses count as part of their matrix clause in German, whereas not fully integrated or non-integrated clauses have independent prosodic contours which indicates that they are not part of their matrix clause (see e.g. Reis 1997). The position of the non-finite part of the verb form counts as end signal for the clause (e.g. Auer 1991:141f.; Uhmann 1993:336f.).

[2] On givenness and other information structural concepts s. Féry & Krifka (2008). On the impact of structuring of information with respect to subordination in general e.g. Fabricius-Hansen 1998, Holler 2009. On adverbial, esp. causal clauses e.g. Redder 1990, Antomo 2012. On finite versus non-finite clauses e.g. Smirnova (2015). On relative clauses e.g. Speyer (to appear). On postfield positioning e.g. Hoberg (1997).

[3] This idea is in line with Hawkins' (2009) principles of *Minimize Domain* and *Minimize Forms*.

Another avenue is taken in computational linguistic surroundings. Based on work by Shannon (1948), the amount of information of a linguistic unit is related to its probability. Specifically, the information. or surprisal (Hale 2001), of a linguistic unit is exactly quantifiable by the likelihood of its appearance (=P) in a given context. Surprisal is calculated by the following formula (Levy et al 2012:15):

$$\text{Surprisal(unit)} = \log_2 (1 / P(\text{unit}|\text{context}))$$

This probability-based notion of information is directly linked to processing effort, which is lower for relatively predictable expressions (Hale 2001).

In order to put this idea into practice and to measure the predictability of a word in context, it is necessary to determine what exactly the context to be considered looks like. Theoretically this is not a trivial issue, as listeners use as diverse sources of information to predict upcoming words as local syntactic context (Jaeger 2010) and script knowledge (Hare et al. 2009). In practice though, it is almost impossible to build a model considering all potential sources of information and weighing them adequately. Therefore, the predictability of a word in context is frequently computed with n-gram language models. Language models are trained on a corpus and return the probability of a word to occur in this corpus in the context of the preceding word(s).

The most simple form of language models are unigram models, which consider no local context but are based on the frequency count of words in the corpus as a whole. Bigram language models take the immediately preceding word as context and thus return the conditional probability of a word on the immediately preceding one, trigram language models are based on a context of two words and so forth. In most cases, n-gram surprisal collapses syntactic and semantic information, for instance, a transitive verb is most probably followed by a noun phrase headed by an article, but the likelihood of specific nouns will also vary in function of the verb.

Language producers strive to distribute the information as evenly as possible. This is a consequence of Shannon's (1948) channel capacity, which shall be approached in order to communicate efficiently but not exceeded, as transgression of channel capacity is penalized with additional processing effort. This has been captured as the Uniform Information Density Hypothesis (UID, Genzel & Charniak 2002, Levy & Jaeger 2007, Jaeger 2010, Crocker et al 2016, a.o.). Applied to the choice of variants, the UID takes care that 'within the bounds defined by grammar, speakers prefer utterances that distribute information uniformly across the signal (information density). Where speakers have a choice between several variants to encode their message, they prefer the variant with more uniform information density (ceteris paribus).' (Jaeger 2010:25).

In real life, we expect some relation to hold between information structural properties such as givenness or topichood – properties that are responsible for high salience of an expression – and surprisal in an information theoretical sense. If this is so we expect the variants presented in section 1.1 to differ both with respect to information structural properties and with respect to surprisal. If this is true, the link between surprisal and properties such as givenness could be strengthened by adding accomodation (in the sense that referents or predicates are 'new' in the discourse universe when they are first introduced, but are 'given' when they are subsequently mentioned) into the model. As the available language models do not take the accomodation of referents in the process of a discourse into account (cf. e.g. Genzel & Charniak 2002), it would be desirable to integrate this into the model.

In this paper we test the assumption that the choice between outsourcing and bundling variants depends on surprisal and is thus influenced by the UID. The test case are relative clauses, the bundling variant being relative clauses in-situ, adjacent to the head noun, the outsourcing variant being clearly extraposed relative clauses. We test this assumption on historic texts, two sermons from the 17$^{th}$ century. Eventually a diachronic study is planned in which these results are corroborated by analysing more data and in which the results are compared to results of later and earlier texts of the same text type. This paper reports the results of a pilot study that shall lay the grounds for a larger study in the context of the 2$^{nd}$ phase of the Collaborative Research Center 1102 'Information Density and Linguistic Encoding' at Saarland University.

## 2 Design of the Pilot Study

### 2.1 The Material

In the study for the CRC 1102, texts of two text sorts shall be compared in some diachronic depth that both have an interesting relation of orality and scripturality (to these concepts s. Koch & Oesterreicher 2007), namely scientific papers and sermons. Scientific papers are a relatively young text type: in German, the earliest examples date to the 2$^{nd}$ half of the 18$^{th}$ century, in English, the writing of academic papers starts around 1665 (Kermes et al 2016). The text type stems from the text type 'letter', a close-to-oral text type. This allows us to observe the evolvement of the text type conventions. We especially expect oral traits (to which the respecting of production constraints such as the UID belongs) to decrease during the history of the text type. Sermon is a text type for which the diachronic depth is large (the earliest German examples are from the 13th century AD). Furthermore, the texts are designed to be read aloud, so we expect oral trait to play a role throughout the time. If there are any changes that do not pertain to the evolvement of the text type but to real change processes, they should be found in such a text type.

For the pilot study we used a subcorpus of the Deutsches Textarchiv (German text archive: DTA). The DTA is a diachronic corpus of different text types of German from the 17th to the 19th century, which is available online (http://www.deutsches-textarchiv.de). Currently (March 3, 2017) it offers digital versions of 2610 works, comprising over 145 million words. It contains texts of different registers and text types, e.g. scientific writing, newspaper text, fiction and function literature. Academic papers in the strict sense are not part of the DTA, but other scientific treatises written in German. As for sermons, there is an abundance of material of a subgenre of sermons, namely funeral sermons. The digital versions are presented as facsimilia with transcription, and are downloadable in several versions. The corpus is available in XML format which includes segmentation into sentences and annotation of lemmata and spelling errors for each word. We extracted the lemmatized version of the corpus from the original files and we modified the original segmentation into sentences by introducting additional sentence boundaries at all periods.

For the current investigation we used the complete funeral sermon section dating from 1600 to 1630, which contains 85.531 sentences (2.03M words). The choice of the time range was arbitrary. The subcorpus used here can later serve as a pivot for diachronic studies as point of comparison to earlier and later texts.

## 2.2   The Realization of the Pilot Study

**The Material**

The main purpose of the pilot study was to see whether surprisal shows an effect at all on the choice of the variants. A second purpose was to test whether accomodation can be included into the model and what its effect is compared to the bare surprisal values. In order to compute the surprisal of individual words in context, we calculated a bigram language model with Kneser-Ney smoothing with the SRILM toolkit (Stolcke 2002) on the whole data set. The vocabulary size of the language model is 77.428. We transformed the language model output surprisal values, $\log_{10}$ p(word|context) into surprisal = $-\log_2$ p(word|context) with a Python script.

In four arbitrarily chosen texts from the corpus (Albinus, Christoph: *Trost Trawriger Eltern*, Brieg 1628; Barthisius, Henoch: *Geistlicher SterbeKittel gleubiger Christenhertzen*, Leipzig 1614; Briaeus, Franciscus: *Leichpredigt / Bey der Begrebnuß*, Marburg 1616; Zuckwolf, Johann Jacob: *Christlich Leichpredigt*, Darmstadt 1614) all relative clauses were identified manually and coded whether the relative clause is unambiguously adjacent to its head noun (the bundling variant, 5a) or unambiguously separated from its head noun and extraposed (the outsourcing variant, 5b). Cases in which one could not tell whether the relative clause is adjacent or extraposed were not taken into consideration. Likewise, we left out cases in which the head noun was extraposed together with its relative clause (5c), as in these cases two conditions are mixed up, postfield positioning and relative clause extraposition.

(5) a.   wer wolte sich denn vber den tödtlichen Abgang
who wanted himself ptc. over the deadly exit
der seinigen / die selig im HErrn sterben /
of-the belonging-to-him who blissfully in-the Lord die
so hoch bekümmern
so highly chagrin
'Who wanted to chagrin so severely about the death of his relatives who die blissfully in the Lord.'   (Barthisius p.20, l.25ff.)
b.   vnd dem HErrn auffzuheben geben / der es verwahren
and to-the Lord to preserve give who it store
vnd wiedergeben kan.
and give-back can
'and to give it to the Lord for preserving it, who can store it and give it back.'
(Barthisius p.17, l.3f.)
c.   Ist also Rahel mit warer Gottseligkeit gezieret gewesen /
is thus Rahel with true beatitude-in-God adorned been
in krafft welcher sie hat können sagen
in power of-which she had can say
'Thus, Rahel was adorned with true beatitude in God in power of which she could tell.'
(Briaeus p. 16, l.1f.)

Relative clauses are the ideal test case for two reasons. First, they tend to be relatively short. Second, and more importantly, there are no interpretatory effects connected with their positioning (and thus the grade of integration). For adverbial clauses, especially causal clauses, there is an ongoing discussion whether the degree of integration is correlated to interpretatory effects.[4] In the main study this will be controllable, but for the pilot study this would be an extra effort.

In total there were 16 bundling variants (that is: clauses containing in-situ relative clauses) and 18 outsourcing variants (that is: clauses with an extraposed relative clause at their right edge). This is, of course, too small a sample to draw valid conclusions, but for the purposes of the pilot study, which only serves to test the methodology of the larger CRC study, this is sufficient.

With these variants, three calculations were made.

**The First Calculation: Givenness.**

The first calculation aims at quantifying the information structural notion of givenness. For each relative clause and each matrix clause, the portion of given referents is calculated. The method was applied already in Speyer (2015, to appear) and is described there in more detail. The referents are categorized in five degrees of givenness / salience: discourse-new referents; new, but inferable referents; given referents which are not salient (which is defined in these studies as referents for which hold that between the present and the last mention intervene more than 10 referents); given referents which are salient (last mention with not more than 10 referents intervening); given referents which are salient and which at the same time constitute the aboutness-topic of the clause.[5] In Speyer (to appear) it was shown that the ratio of given referents influences the choice between in-situ and extraposed relative clauses.

**The Second Calculation: Bare Surprisal Values.**

The second calculation pertains purely to the surprisal values of the words in the relative and matrix clauses. Here, the surprisal values of a bi-gram language model were collected for each lemma in the relative and matrix clauses and were summed up (additive surprisal value, adS). From this sum the arithmetic mean was calculated (average surprisal value, avS). The first word of each clause was left out, in the case of the relative clauses because there is no relevant variation to be expected: In German, relative clauses are always introduced by a relative pronoun.[6] There are two kinds of relative pronouns, *der/die/das* and *welcher/-e/-es*. The choice between them depends on stylistic considerations, and presumably not on information density. Thus, the inclusion of the relative pronouns would add irrelevant noise. For reasons of symmetry, the first words of the matrix clauses have been left out as well. We chose the type-based lemma level instead of the token-based word level, as we are not interested in the occurrence of specific word forms, but on the occurrence of lexical items per se. Furthermore, as language modeling operates on word forms in the training data, this reduces the vocabulary size and returns more meaningful probabilities given the relatively small data set.

A crucial difference to the first calculation is that in the second calculation the unit of investigation is words, more concretely: lemmata, and not referents. The reason for this is that the corpus we used is lemmatized and thus the language models can deal with words or lemmata without any problems. To integrate referents into the language model, the corpus would have to be annotated for referents (and not only the few texts on which the calculations were made but the whole training corpus), which is very time-consuming, as it can hardly be automatized.

**The Third Calculation: Surprisal with Accomodation.**

The third calculation strives to add accomodation to the bi-gram-based language model. The basic idea is that any lexical item should come with high 'surprisal' in the non-technical sense of the word when it is used in a text for the first time, simply because it cannot be part of the discourse universe up to the moment it is introduced into the discourse. This 'novelty bonus' should wear out after a while, and after it has been mentioned a certain number of times, it should make no difference any more. We try to model this by multiplying the surprisal values of the second calculation with the y values of some hyperbolic function, the x values being the serial number of occurrence of the lexical item under consideration. Crucial are two values: the y value with $x = 1$ (which would be the 'novelty bonus'), and the x value with $y = 1$ (which would be the point when the novelty bonus wears out). These values should have some bearing on measurable cognitive effort, which we will try to elicit in experiments in the CRC project. For the moment, we chose an arbitrary hyperbolic function, namely:

$$y = 4/x$$

---

[4] Non-integrated causal clauses have been claimed to trigger a 'mediate', non-factual interpretation, either giving a reason for the assumption underlying the main clause (epistemic usage) or the illocution of the main clause (speech act related usage). The literature is cited in Antomo & Steinbach (2010). Especially working with spoken and diachronic data, it becomes clear, however, that these effects do not hinge on the degree of integration, but that factual causal clauses can be non-integrated as well (Scheutz 1998, Speyer 2011).

[5] Although there is a certain typical intersection between (aboutness) topics and given information, these concepts should not be equated, as it has been done in research on information structure in the 1970s up to the 1990s. See Féry & Krifka (2008).

[6] Relative particles comparable to the English *that* occur every now and then (*so* in Early New High German, *wo* in Modern German dialects). In German, there are no non-introduced relative clauses of the type *The man I saw yesterday returned to the shop*. In English, the variation between introduced and non-introduced relative clauses have been found to be influenced by information density (e.g. Jaeger 2010).

This means that the novelty bonus is set at 4 times the 'normal' effort which is represented by the surprisal value. The point after which the bonus wears out is the fourth mentioning of the concept. The point at the moment is simply to look whether introducing accomodation into the model makes any difference, so it is excusable if an arbitrary function (which is also easily calculated) is chosen. If some effect appears, we interpret this as indication that it is worthwhile to hunt for a more realistic function.

Accomodation, however, does not only mean that any concepts (referents or predicates) are new at some stage of the text and more easily accessible after that. What happens if a concept is introduced, dealt with for a certain portion of the text, then not mentioned for a while and eventually re-introduced into the discourse? The re-introduction cannot count as new information; on the other hand, the concept is presumably not in the working memory of the participants, if it is activated at all after it has been abandoned for a certain time span. We try to bring this property into the model by a resetting mechanism: Once a lexical item has not been used for 200 words (which corresponds roughly to one page in the texts that we used), its x value is reset 1 point for every 200 words. The reset value cannot fall below $x = 2$, however, in order to keep the absolute 'novelty bonus'. Let us illustrate this with an example. A certain word appears for the first time as, say, the $624^{th}$ word in the text. This is $x = 1$. By the function $y = 4/x$ it will receive the factor 4, which means that its surprisal value is multiplied by 4. The word appears again as the $656^{th}$ word ($x = 2$), it receives the factor 2. The third occurrence is as the 681th word ($x = 3$, factor 4/3), the fourth occurrence is as the $702^{nd}$ word ($x = 4$, factor 1) and the fifth as the $715^{th}$ word ($x = 5$, factor 1; for $x > 4$ we would set the y value at 1). Then the word would not be used for a certain time, but reappear only as the $1267^{th}$ word. Between the $5^{th}$ and the $6^{th}$ mentioning there are 564 words intervening. This means, the $6^{th}$ mentioning is treated as if it were the $3^{rd}$ mentioning: At word no. 915, the x value is reset to 4, at word no. 1115, the x value is reset to 3. So its surprisal value is multiplied by the factor 4/3. Then the word is not reused until word 2785. By the resetting algorithm, this should mean that the x value is treated as if it were -4. As the threshold for resetting is set at $x = 2$, it is treated as if it was the second occurrence and its surprisal value multiplied by 2.

## 3   The Results

### 3.1   First Calculation (Givenness)

This is a very rough measure, and because of the small size of the data set it does not reach the level of statistic significance ($\chi^2$-test for new referents: $p = 0.1459$). However, the tendency is the same as in the relative clauses reported in Speyer (to appear), as shown in Table 1.

**Table 1.** Ratio of new and salient referents.

|  | referents total | new referents | salient referents |
|---|---|---|---|
| Relative clauses: in-situ | 22 | 2<br>9% | 22<br>55% |
| Matrix clauses of in-situ rel. cl. | 40 | 4<br>10% | 24<br>60% |
| Relative clauses: extraposed | 46 | 11<br>24% | 19<br>41% |
| Matrix clauses of extraposed rel.cl. | 39 | 14<br>36% | 23<br>59% |

The ratio of salient given referents is higher in the relative clauses that stand in-situ than in the ones that are extraposed. Likewise, the ratio of new referents is lower in the in-situ relative clauses, but at 24% in the extraposed relative clauses. Note that the matrix clauses of in-situ relative clauses show also a low ratio of new referents, compared to the matrix clauses of extraposed relative clauses. We might interpret this in such a way that clause complexes which do not contain many new referents, neither in the relative nor their matrix clause, allow the bundling variant more freely.

### 3.2 Second Calculation (Bare Surprisal Values)

The results of the second calculation are given in Table 2. Here, the surprisal values of all items (= lemmata), calculated with a bi-gram language model, were added (adS) and the mean was calculated (avS). This was done for all clauses (relative clauses and there matrix clauses) separately, in the case of in-situ relative clauses, the adS and avS of the combined clause was calculated, too. Note that this is not simply the sum of the values of the in-situ relative clause and its matrix clause. For the surprisal values of the clauses by themselves, in-situ relative clauses were calculated as if they were extraposed. As the surprisal values are calculated via bigrams, the context of some words would change whether the relative clause is in-situ or not, most notably the surprisal value of the word immediately following the in-situ relative clause.

**Table 2.** Additive and average surprisal values of lemmata per clause

| extraposed (=outsourced) relative cl. | | adS | avS |
|---|---|---|---|
| n =18 | rel. cl. | 71.204 | 8.3766 |
| | matrix cl. | 59.2219 | 7.6952 |
| in-situ (= bundled) relative cl. | | | |
| n = 16 | rel. cl. | 39.1147 | 7.2166 |
| | matrix cl. | 65.901 | 7.6386 |
| | combined | 112.4235 | 7.4613 |

The difference of avS between the matrix clauses of in-situ relative clauses and the matrix clause of extraposed relative clauses is negligible. The relative clauses themselves, however, show some difference: the avS of the in-situ relative clauses is with roughly 7.2 not only smaller than the avS of the matrix clauses (7.6 to 7.7), but also conspicuously smaller than the avS of the extraposed relative clauses (8.4). The avS of in-situ relative clauses is roughly 86% of the avS of extraposed clauses, for avS this is a relatively big deviation. Looking at the adS of in-situ relative clauses, we see that it is very small in comparison, only 59% of the adS of their matrix clauses and, more importantly, only 55% of the adS of extraposed relative clauses. The adS of extraposed relative clauses is slightly higher than the adS of their matrix clauses (the adS of the matrix clauses is 83% of the adS of extraposed relative clauses), but on the whole the adS of the matrix clauses and the extraposed relative clauses are similar.

The bundled variant shows an avS of roughly 7.5 which is close to the avS of the matrix clauses.

### 3.3 Third Calculation (Surprisal with Accomodation)

The results of the third calculation are given in Table 3. Here, the surprisal value of each content word was multiplied by a factor which was determined using the algorithm described in section 2. The adS and avS were calculated subsequently as in the second calculation.

**Table 3.** Additive and average surprisal values with factored in accomodation

| extraposed (=outsourced) relative cl. | | adS | avS |
|---|---|---|---|
| n =18 | rel. cl. | 151.81 | 18.3556 |
| | matrix cl. | 114.8447 | 15.4172 |
| in-situ (= bundled) relative cl. | | | |
| n = 16 | rel. cl. | 77.1524 | 13.8858 |
| | matrix cl. | 114.1298 | 13.9899 |
| | combined | 197.2565 | 13.184 |

Similar effects are visible to the bare surprisal values of the second calculation, in that there is a striking difference between the values of extraposed and in-situ relative clauses. The avS of the extraposed variants is roughly 18.4, much higher than the avS of the in-situ variants (13.9) which is 76% of the extraposed value. A difference however is that the accomodated surprisal values of the matrix clauses matter, too: The avD of matrix clauses to in-situ relative clauses is almost identical to the avD of their relative-clauses. The adS values, on the other hand, pattern more with what we would expect from the second calculation.

# 4    Discussion

The surprisal values are perceived as an approximative measurement of cognitive effort (see e.g. Hale 2001, Crocker et al 2016). As such, the low adS value of in-situ relative clauses by themselves indicates that they are very easy to process, at any rate conspicuously easier than their extraposed counterparts. That the adS values of matrix clauses to both kinds of relative clauses are similar, and also similar to the ones of extraposed relative clauses, supports the Uniform Information Density Hypothesis (UID; e.g. Genzel & Charniak 2002, Levy & Jaeger 2007, Jaeger 2010, Crocker et al 2016): The actual realizations of these clauses are such that relative clauses with a relatively high adS are extraposed and thus removed from the clause proper so they can be processed separately. In principle, processing can be understood on behalf of both the producer and the perceiver, here it is understood from the perspective of the producer: The producer, after noticing that the content of the relative clause is likely to become difficult to process, opts for the extraposition variant quite early in planning.

The UID is even more visibly at play if we look at the avS values. The avS value of in-situ relative clauses by themselves is lower than the value of the actually produced clauses in which in-situ clauses are an integral part of their matrix clause and have to be processed together. The avS value 7.4613 of bundled clauses is not so far from the avS value 7.6952 of matrix clauses of extraposed relative clauses, which can be seen as an average value of a normal clause. This can be interpreted such that bundling is used as a strategy to avoid troughs in the information density profile which would arise if these relative clauses were extraposed and thus processed separately from their respective matrix clauses.

As to the accomodated surprisal values, the fact that the values of matrix and relative clauses are nearly identical in case of in-situ relative clauses might indicate that the information put forward in these clause complexes is relatively low anyway, so bundling does not do any harm and will not surpass the channel capacity. Looking at the adS values, we see the matrix clauses of both types patterning together, the in-situ relative clause showing a very low figure, less than half the adS value of extraposed relative clauses. While the adS might not say very much about the distribution of information – for this the avS value is surely the more adequate measurement – it might say something about the capacity, that is, how much information can be processed simultaneously. I might be interesting to note that we calculated the hypothetical surprisal values of the extraposed relative clauses as if they were in-situ: while the avS was not very different from the respective matrix clauses (bare surprisal values: 7.9738, accomodated values: 15.3488), the adS values were of course much higher (bare surprisal values: 130.9256, accomodated values: 257.4992). Most notably, the adS values of these hypothetical bundled cases were much higher than the adS values of the 'real' bundling cases (bare values: 130.9256 compared to 112.4235, which is 86%; acc accomodated values: 257.4992 compared to 197.2565, which is 77%), so the lower values of the 'real' clause complexes (the attested in-situ relative clauses) might reflect a value below the upper boundary of the channel, the higher value of the 'hypothetical' clause complexes (the attested extraposed cases which are mashed into their matrix clauses) might reflect a value that is above the threshold.

All three calculation point in the same direction and show a difference between in-situ relative clauses and extraposed relative clauses, with respect to the givenness of their respective referents, the surprisal of the words used in them respectively and a combination of both. The fact that the first and the second calculation point in the same direction, although completely different things are measured (the salience of a referent versus the likelihood of appearance of one word A after another word B) can be interpreted to indicate that there is an intrinsic relationship between these two parameters. If this is on the right track, studying givenness profiles in a text can be used as a proxy to information density (as was tacitly assumed in e.g. Speyer 2015, 2016, to appear), even in a technical sense, while calculating surprisal values can feed into modeling the attention states of text producers.

The text type which we looked at (funeral sermons) was chosen because it has several advantages. First, it is a close-to-oral genre in that it is produced for being read out loud. Thus, in order to be understood, the writer / preacher must take the constraints of on-line production into consideration and produce his sentences accordingly. Second, it is a very homogenous genre with respect to the situation to which it pertains and the expectations that are connected with it. The expectations can be mirrored by using a training corpus of exactly this genre, so one factor that has been found to influence the expectations, namely the field of frame and script knowledge (see Crocker et al 2016), is controlled. Third, examples for such sermons cover a wide range of the German language history, so in studying this text type (and by that controlling this factor) we have a chance to see whether any changes occurred. As it is necessarily a close-to-oral text type, stylistic developments, as they are for instance found in fictional literature, or the evolvement of a text type, as is the case with e.g. the text type of academic papers, should play a minor role. If we witness some changes in a diachronic study, to which the present pilot study serves as a pivot, these changes are probably 'real' and may shed light on the influence of factors such as information density on language change.